\documentclass[11pt]{article}
\usepackage[final]{acl}

\usepackage{times}
\usepackage{latexsym}

\usepackage[T1]{fontenc}

\usepackage[utf8]{inputenc}

\usepackage{microtype}

\usepackage{inconsolata}

\usepackage{graphicx}
\usepackage{booktabs}

\usepackage{amsmath}
\usepackage{xcolor}  
\newcommand{\zr}[1]{\textcolor{black}{#1}}
\newcommand{\benchmark}{\textsc{MultiTextEdit}}

\newcommand{\best}[1]{\textbf{#1}}
\newcommand{\second}[1]{\underline{#1}}
\newcommand{\affmark}[1]{\textsuperscript{\ensuremath{#1}}}

\title{\benchmark{}: Benchmarking Cross-Lingual Degradation in Text-in-Image Editing}

\author{
\begin{tabular}{@{}c@{}}
Liwei Cheng\affmark{1,*} \quad
Shibo Feng\affmark{1} \\
Lunjie Zhou\affmark{1} \quad
Yixuan Guan\affmark{1} \quad
Dayan Guan\affmark{1,\dagger} \\[0.6em]
\affmark{1}Harbin Institute of Technology \\
{\small \affmark{*}Equal contribution. \quad \affmark{\dagger}Corresponding author.}
\end{tabular}
}

\begin{document}
\maketitle
\begin{abstract}

\end{abstract}
\zr{Text-in-image editing has become a key capability for visual content creation, yet existing benchmarks remain overwhelmingly English-centric and often conflate visual plausibility with semantic correctness. We introduce MULTITEXTEDIT, a controlled benchmark of 3,600 instances spanning 12 typologically diverse languages, 5 visual domains, and 7 editing operations. Language variants of each instance share a common visual base and are paired with a human-edited reference and region masks, isolating the language variable for cross-lingual comparison. To capture script-level errors that coarse text-matching metrics miss, such as missing diacritics, reversed RTL order, and mixed-script renderings, we introduce a language fidelity (LSF) metric scored by a two-stage LVM protocol that first traces the edited target text and then judges it in isolation, reaching a quadratic-weighted $\kappa$  of 0.76 against native-speaker annotators. Evaluating 12 open-source and proprietary systems with LSF alongside standard semantic and mask-aware pixel metrics, we find pronounced cross-lingual degradation for every model, largest on Hebrew and Arabic and smallest on Dutch and Spanish, and concentrated in text accuracy and script fidelity rather than in coarse structural dimensions. We also uncover a pervasive semantic and pixel mismatch, where outputs preserve global layout and background fidelity yet distort script-specific forms.}

\section{Introduction}

Text-in-image editing aims to modify text rendered in an image while preserving the surrounding visual content.
Recent multimodal and diffusion-based systems can replace, insert, delete, recolor, and reposition in-image text with increasingly strong visual quality~\citep{tuo2023anytext, tuo2024anytext2, liu2024textmastero, kong2025fluxtext}.
This capability is important for graphic design, advertisement localization, event promotion, and product marketing, where the same visual asset often needs to be adapted across languages.
Unlike generic image editing, text in images is an editable object that carries both visual form and linguistic meaning.
This dual nature becomes especially challenging in multilingual settings, where writing systems differ in typography, character composition, spacing conventions, and writing direction.

Multilingual evaluation is therefore essential.
Real-world content is edited not only in English, but across languages that differ widely in script and resource availability, especially in non-Latin and lower-resource settings.
Yet current evaluation remains largely English-centric~\citep{zhu2024textediibench, wu2024vtpbench}.
Beyond this coverage issue, existing evaluation protocols often conflate visual plausibility with semantic correctness.
For text-in-image editing, an output may look locally consistent while still rendering the wrong word, failing to follow the instruction, or distorting script-specific characters.
As a result, we still lack a controlled way to measure how strongly state-of-the-art multimodal models degrade across languages.

To address these issues, we introduce \benchmark{} (\textbf{Multi}lingual \textbf{Text} \textbf{Edit}ing Benchmark), a benchmark for analyzing multilingual degradation in text-in-image editing.
\benchmark{} contains 3,600 instances created from 300 base images and their 12 language variants, covering 5 visual domains and 7 editing operations.
Each instance includes a source image, an editing prompt, a human-edited ground-truth image, an operation label, a domain label, and a mask for the edited region.
The benchmark spans Latin, CJK, Arabic, Hebrew, Cyrillic, and Bengali scripts, enabling controlled analysis across diverse writing systems and language-resource conditions.
Because different language versions share the same underlying visual content and edit intent, the benchmark supports controlled cross-lingual comparison rather than loosely matched multilingual evaluation.

We further adopt a dual-track evaluation protocol that combines semantic assessment with pixel-level fidelity.
The semantic track measures instruction following, text accuracy, visual consistency, and layout preservation, while the pixel track uses standard similarity metrics to quantify closeness to the reference image.
This design lets us study not only overall performance, but also the \emph{semantic-pixel mismatch} that arises when visually plausible edits are semantically wrong.
Our experiments on several strong multimodal models reveal substantial performance degradation outside English together with recurring mismatches between semantic success and pixel similarity.

Our contributions are threefold:
(1) we introduce MULTITEXTEDIT, a controlled multilingual benchmark for text-in-image editing across 12 languages, 5 domains, and 7 editing operations; (2) we introduce a language/script fidelity (LSF) metric with a two-stage tracing-and-scoring protocol that diagnoses script-level errors current metrics miss, validated by a $\kappa=0.76$ agreement with native-speaker annotators; (3) we provide a systematic empirical study across 12 models revealing pronounced cross-lingual degradation and a recurring semantic–pixel mismatch.

\section{Related Work}
\label{sec:related}

\paragraph{Text-in-Image Editing Models.}
Diffusion-based and multimodal methods have rapidly advanced controllable text rendering and editing, including the TextDiffuser series~\citep{chen2023textdiffuser, chen2024textdiffuser2}, the AnyText series~\citep{tuo2023anytext, tuo2024anytext2}, and more recent work targeting disentangled editing, recognition-aware consistency, glyph fidelity, and low-resource language support~\citep{zhang2024choose, fang2025recognition, wang2025glyphmastero, stellar2024}.
While these models substantially improve visual quality and controllability, they do not by themselves characterize how performance changes across languages, scripts, and resource conditions, motivating a benchmark focused on multilingual robustness.

\paragraph{Benchmarks for Text Editing and Multilingual Visual Text Tasks.}
Existing benchmarks evaluate text-related editing or visual text processing but leave the multilingual degradation question open. TextEditBench~\citep{zhu2024textediibench} targets text-in-image editing without controlled cross-lingual comparison; VTPBench~\citep{wu2024vtpbench} covers a broad range of visual text tasks; ScenePair~\citep{zeng2024textctrl} provides paired supervision for scene text editing; and IMTBench~\citep{imtbench2025} studies in-image machine translation rather than general editing operations. Scene text recognition research has also long highlighted script-dependent difficulty~\citep{shi2016end, baek2019what, bautista2022scene}. In contrast, \benchmark{} is built specifically to compare the same editing intent across 12 languages under shared visual contexts, enabling direct analysis of cross-lingual performance drop.

\paragraph{Evaluation of Text-in-Image Editing.}
Pixel and perceptual metrics such as SSIM and LPIPS measure structural and perceptual similarity~\citep{wang2004image, zhang2018unreasonable} but cannot determine whether the edited text satisfies the instruction. Recent image editing evaluation therefore emphasizes instruction faithfulness alongside visual fidelity~\citep{qian2025gie, wu2025kris}, and LLM-as-a-judge paradigms~\citep{zheng2023judging} provide a scalable way to score open-ended generation. Our framework builds on these insights but tailors them to multilingual text-in-image editing, where correctness must be judged at both the semantic and visual/layout level, and where cross-language comparison is a central goal.

\begin{figure*}[t]
\centering
\includegraphics[width=\textwidth, trim=0.7cm 3.5cm 0.5cm 2.8cm, clip]{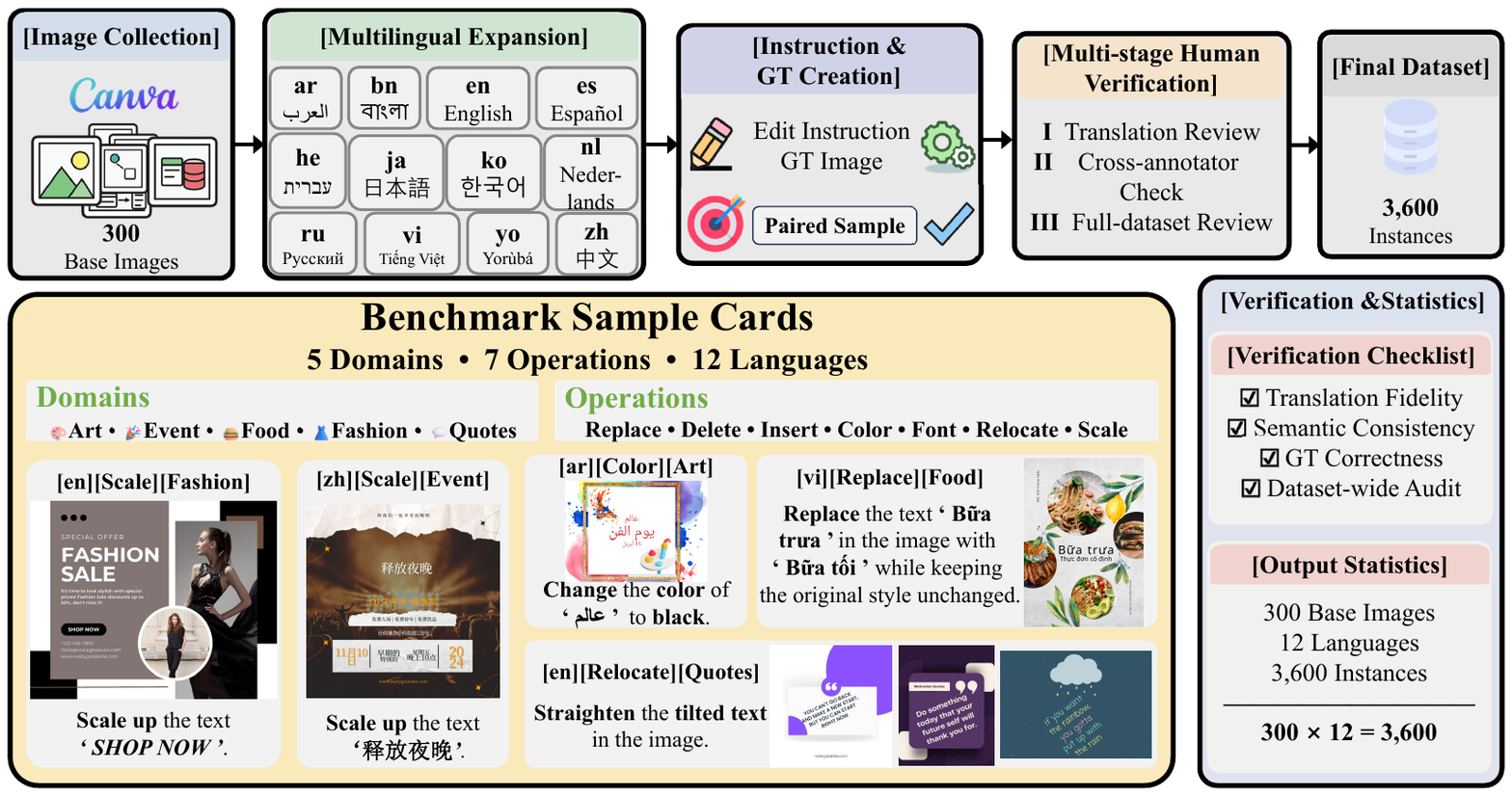}
\caption{Overview of the \benchmark{} data construction pipeline.}
\label{fig:benchmark-pipeline}
\end{figure*}

\section{\benchmark{}}
\zr{We present MULTITEXTEDIT, a multilingual benchmark for text-in-image editing evaluation. The benchmark is constructed by extending 300 base visual samples across 12 languages, yielding 3,600 annotated instances with paired input-output images, editing instructions, and masks.}

\subsection{Task Definition}

\zr{Given a source image $I$ containing text in a target language and an editing instruction $p$, the task is to produce an edited image $\hat{I}$ that applies the specified operation to the designated text region while preserving the remaining visual content. Our focus is not on multilingual instruction understanding; we therefore keep $p$ in English across all language variants of a base image and vary only the target text content. This design isolates the language of the rendered text as the single controlled variable across comparable samples.}

\zr{Formally, each instance is an 8-tuple $x = (I,\ p,\ I^*,\ M_{\text{src}},\ M_{\text{tgt}},\ l,\ o,\ d)$, where $I$ is the input image, $p$ the editing instruction, $I^*$ the human-edited reference, $M_{\text{src}}$ and $M_{\text{tgt}}$ binary masks for the source and target text regions, and $l \in \mathcal{L}$, $o \in \mathcal{O}$, $d \in \mathcal{D}$ the language, operation, and domain labels respectively. The dual masks support region-aware evaluation at both edited and non-edited regions.}

\subsection{Benchmark Construction}
\paragraph{Base images} We source 300 base images from Canva poster templates\citep{canva2025about}, which provide diverse layouts, fonts, and visual styles suitable for controlled text editing.
A larger candidate pool was automatically crawled and then manually filtered for legibility, absence of visual artifacts, and sufficient editable space around text regions. 
We summarize the full construction pipeline in Figure~\ref{fig:benchmark-pipeline} .

\zr{\paragraph{Multilingual expansion}Each base image is expanded into 12 language versions which include Arabic, Bengali, Chinese, Dutch, English, Hebrew, Japanese, Korean, Russian, Spanish, Vietnamese, Yoruba by translating only the in-image text. Compared with collecting native-language images independently per language, this shared-base design controls for background, composition, font style, and overall visual context, so that cross-lingual differences in model output can be attributed primarily to the target script and language rather than to confounding visual variation.}

\zr{\paragraph{Editing instructions} For each base image, we manually write an English editing instruction following a fixed template per operation type. The same template is reused across the 12 language variants, with only the quoted target text substituted, so that instruction difficulty is held constant across languages.}

\paragraph{Data quality} Annotation was carried out by a team of international students who are native speakers of the target languages, under a three-stage protocol. (i) Translation review: initial translations produced by GPT-5.2 ~\citep{openai2025gpt52} were proofread by the corresponding native speaker, who corrected translation errors and flagged cases where the translated text conflicted with the original layout, caused occlusion, or produced visually implausible arrangements. (ii) Reference editing and cross-check: human experts manually edited each image in Canva to produce the ground-truth reference, which was then reviewed by a second annotator for text correctness, instruction–operation consistency, and visual naturalness of the edited region. (iii) Dataset-wide audit: a final pass over the full dataset checked for residual text occlusion, instruction errors, and reference–annotation mismatches.

\zr{\paragraph{Domain distribution} The benchmark covers five visual domains (\texttt{Art}, \texttt{Event}, \texttt{Fashion}, \texttt{Food}, \texttt{Quotes}) and seven editing operations; their joint distribution is shown in Table~\ref{tab:benchmark-stats}. We preserve the naturally imbalanced distribution rather than artificially balancing it, as it better reflects real-world design scenarios, and report stratified results by domain and operation in our experiments.}

\begin{table}[t]
\centering
\small
\setlength{\tabcolsep}{4pt}
\caption{Distribution of base samples across editing operations and visual domains in \benchmark{}. Each base sample is expanded into 12 language versions.}
\label{tab:benchmark-stats}
\begin{tabular}{l ccccc c}
\toprule
\textbf{Operation} & \textbf{Art} & \textbf{Event} & \textbf{Fashion} & \textbf{Food} & \textbf{Quotes} & \textbf{Total} \\
\midrule
Delete    & 15 & 10 & 30 & 5 & 5 & 65 \\
Color     & 15 & 10 & 20 & 5 & 5 & 55 \\
Replace   & 10 & 10 & 15 & 5 & 5 & 45 \\
Insert    & 10 & 10 & 10 & 5 & 5 & 40 \\
Relocate  & 10 & 10 & 10 & 5 & 5 & 40 \\
Scale     & 10 & 10 & 10 & 5 & 5 & 40 \\
Font      & -- & -- & 15 & -- & -- & 15 \\
\midrule
\textbf{Total}          & 70 & 60 & 110 & 30 & 30 & \textbf{300} \\
\bottomrule
\end{tabular}
\end{table}

\subsection{Language Coverage}

To systematically evaluate multimodal models' text-in-image editing capabilities across diverse linguistic conditions, \benchmark{} covers 12 languages selected according to two principles: (1) resource-level diversity, including both high-resource languages and lower-resource languages; and (2) typological diversity, spanning different writing systems, character composition mechanisms, and text organization patterns. The Meta columns of Table~\ref{tab:lang-degradation} summarize the script, directionality, and resource level of each language.

The selected languages exhibit substantial typological variation~\citep{wu2024vtpbench}.
Latin-script languages (English, Spanish, Dutch, Vietnamese, Yoruba) serve as a baseline group, though Vietnamese and Yoruba introduce additional complexity through diacritics and tonal marks that affect glyph rendering~\citep{lehong2021vietdiacritics, asahiah2017yoruba}.
CJK languages (Chinese, Japanese, Korean) feature high character density, complex stroke patterns, and support for vertical text layout, posing challenges for both recognition and spatial editing~\citep{w3c2026clreq, w3c2020jlreq, w3c2024korelreq}.
Right-to-left scripts (Arabic, Hebrew) reverse the default text flow, requiring models to handle mirrored layout conventions~\citep{w3c2025alreq, w3c2024hebrlreq}.
Russian uses Cyrillic script, which shares structural similarities with Latin but differs in glyph shapes, while Bengali employs an abugida writing system with conjunct consonants and vowel diacritics that form ligatures~\citep{w3c2024benglreq}.
These linguistic differences provide contrastive dimensions for analyzing cross-lingual performance degradation in our experiments.

\section{Evaluation Framework}
We design a dual-track evaluation framework that combines semantic-level assessment with pixel-level metrics, motivated by the observation that neither alone captures the full picture of multilingual text editing quality.
\subsection{Design Rationale}
\label{subsec:design-rationale}

Evaluating text-in-image editing requires judging both whether the target text is edited correctly at the semantic level and whether the non-edited regions remain visually intact. Conventional pixel-level metrics such as SSIM and LPIPS cannot perceive textual meaning, which leads to two characteristic failure modes. First, a semantically correct edit may be over-penalized because of minor deviations in font style, spacing, or position relative to the reference image. Second, a missing diacritic or slight glyph error may be almost invisible at the pixel level while still changing the meaning of the word entirely, such as the Vietnamese word \emph{m\d{e}} (`mother') becoming \emph{me} (`tamarind'). This limitation becomes even more severe in multilingual settings, where languages differ substantially in writing system, character density, word length, directionality, and glyph composition, making cross-lingual evaluation itself more difficult. OCR-based alternatives are also imperfect, especially for scripts and lower-resource languages that are less reliably recognized by current text recognition systems~\citep{shi2016end,baek2019what,bautista2022scene}, so OCR noise can obscure the true editing ability of the model.

\begin{figure*}[t]
\centering
\includegraphics[width=\textwidth, trim=0.15cm 5.45cm 0.05cm 4.35cm, clip]{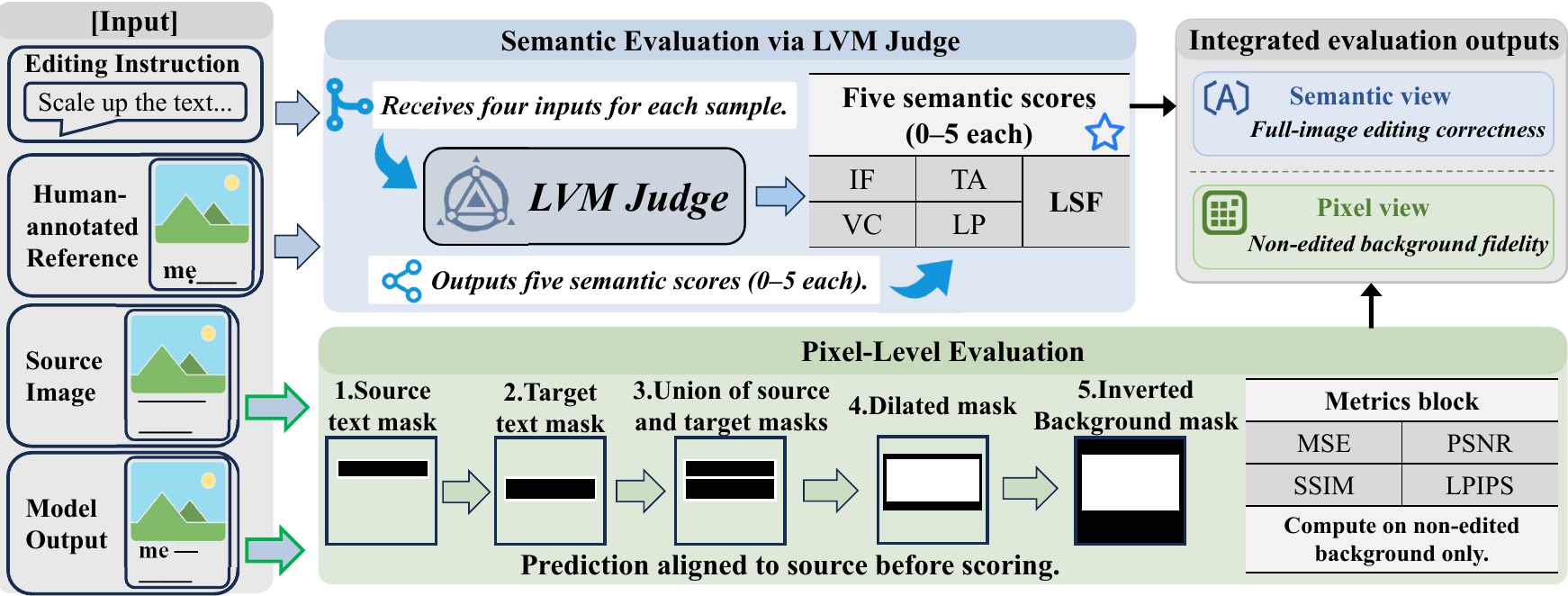}
\caption{Overview of the dual-track evaluation framework. The semantic track uses an LVM judge to score instruction following (IF), text accuracy (TA), visual consistency (VC), layout preservation (LP), and language/script fidelity (LSF) from the source image, model output, reference image, and editing instruction. The pixel track evaluates background preservation on the non-edited region defined by the union and dilation of source and target text masks.}
\label{fig:eval-framework}
\end{figure*}

Motivated by these challenges, we adopt a dual-track evaluation framework. The semantic track uses an LVM judge to assess editing correctness, covering instruction following, text accuracy, visual consistency, layout preservation, and script fidelity (Section~\ref{subsec:semantic-eval}). The pixel track uses mask-aware metrics to separate edited regions from the background and quantify how well non-edited content is preserved (Section~\ref{subsec:pixel-eval}). Following recent model-based automatic evaluation paradigms~\citep{zheng2023judging,qian2025gie,wu2025kris}, we use an LVM judge for scalable semantic evaluation, and we further validate its reliability through a human agreement study in Section~\ref{subsec:human-validation}. Figure~\ref{fig:eval-framework} provides an overview of the full evaluation pipeline.

\subsection{Semantic Evaluation via LVM Judge}
\label{subsec:semantic-eval}

For each sample, the LVM judge receives the source image, the model's edited output, the human-annotated reference, and the editing instruction. It returns an independent integer score on a 0--5 scale for five dimensions. The four general editing dimensions are summarized in Table~\ref{tab:general-dimensions}.

\begin{table}[t]
\centering
\small
\setlength{\tabcolsep}{4pt}
\caption{General semantic dimensions used by the LVM judge.}
\label{tab:general-dimensions}
\begin{tabular}{l p{0.70\columnwidth}}
\toprule
\textbf{Dimension} & \textbf{Definition} \\
\midrule
IF & Whether the model performs exactly the requested operation . \\
TA & Whether the resulting text content matches the target. \\
VC & Whether the edited region blends naturally with the surrounding image . \\
LP & Whether non-target regions remain unchanged. \\
\bottomrule
\end{tabular}
\end{table}

\paragraph{Language/Script Fidelity (LSF).}
The four dimensions above are largely language-agnostic: IF, VC, and LP do not depend on the target language, and TA captures content-level correctness but is insensitive to script-level errors that do not alter the recognizable meaning of a word. However, multilingual text editing introduces a distinct class of failure that TA alone cannot diagnose. A model may produce text whose coarse content is recognizable yet whose script realization is incorrect---missing diacritics in Vietnamese, reversed character order in Arabic, missing tone marks in Yoruba, or mixed scripts within a single word. We therefore introduce LSF to specifically evaluate script-level fidelity, covering character correctness, diacritics and tonal marks, writing directionality (RTL/LTR), and script purity. To illustrate the distinction: if a model renders the Vietnamese target \emph{m\d{e}} as \emph{me}, TA may remain moderately high because the base word is recognizable, whereas LSF should drop substantially because the diacritic is missing.

\begin{figure}[t]
\centering
\includegraphics[width=\columnwidth, trim=2.35cm 4.55cm 16.55cm 4.20cm, clip]{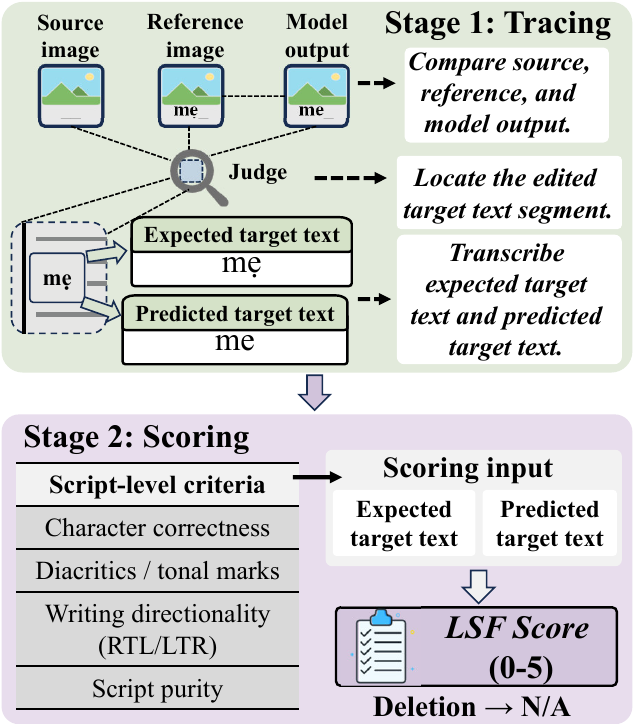}
\caption{Two-stage protocol for language/script fidelity (LSF) scoring. The judge first traces the edited target text by comparing the source image, reference image, and model output, then scores script-level fidelity using only the expected and predicted target text, focusing on character correctness, diacritics or tonal marks, writing directionality, and script purity.}
\label{fig:lsf-two-stage}
\end{figure}

Directly scoring script fidelity on the full image is unreliable because the judge may be distracted by other, unedited text regions. We therefore adopt a two-stage protocol. In the tracing stage, the judge compares the source image and the reference image to identify the target text segment, and transcribes both the expected target text and the corresponding text observed in the model's prediction. In the scoring stage, the judge evaluates script fidelity based solely on the traced target segment, ignoring all other text in the image. This decomposition ensures that LSF focuses exclusively on the edited text and is not confounded by unrelated content. Figure~\ref{fig:lsf-two-stage} illustrates this protocol. For text deletion samples, where no target text is expected in the output, LSF is marked as not applicable.

\subsection{Pixel-Level Evaluation}
\label{subsec:pixel-eval}

While LP in Section~\ref{subsec:semantic-eval} provides a coarse subjective assessment of whether non-target regions are preserved, it cannot precisely quantify unintended pixel-level modification. We therefore complement the semantic metrics with four standard pixel-level metrics---MSE, PSNR, SSIM, and LPIPS---that measure background fidelity between the source image and the model's prediction.

To delineate the editing region, we adopt a dual-mask design. As illustrated in Figure~\ref{fig:eval-framework}, for each sample, annotators manually mark the text region in the source image and the corresponding edited region in the reference image as two separate binary masks. The two masks are merged by taking their union to accommodate edits that alter text position or size, then dilated to buffer against minor spatial misalignment between the model's edit area and the annotation. The dilated mask is inverted to obtain the background region, and all pixel-level scores are computed using this mask so that only non-edited content contributes to the score. Prior to scoring, the prediction is geometrically aligned and resized to the source image.

MSE and PSNR measure low-level pixel agreement, SSIM captures structural similarity, and LPIPS quantifies perceptual distance using deep features.

\subsection{Human Agreement Validation}
\label{subsec:human-validation}

To assess the reliability of the LVM-based evaluation, we conduct a human--LVM agreement study. We sample 20 model predictions per language (240 in total) from the pooled outputs of all evaluated models, stratified to ensure coverage of all editing operation types. Each sample is scored by one native speaker of the corresponding language using the same five dimensions and the same 0--5 scale as the LVM judge. For text deletion samples, LSF is marked as not applicable and excluded from the corresponding agreement computation. We quantify agreement using Quadratic Weighted Cohen's $\kappa$ (QWK) for ordinal agreement and Spearman's $\rho$ for rank-order correlation. In aggregate, the LVM judge achieves an overall QWK of 0.7626 and an overall Spearman's $\rho$ of 0.7951. Agreement remains consistently high across languages, with per-language QWK ranging from 0.6684 to 0.7771 and per-language Spearman's $\rho$ ranging from 0.7049 to 0.8137 .
These results suggest that the LVM judge provides a reliable approximation of human judgment across diverse scripts and resource conditions, supporting its use as the primary semantic evaluator in our benchmark.

\section{Experiments}
\label{sec:experiments}

We evaluate a diverse set of open-source and proprietary image editing systems on \benchmark{}.
All open-source model inference and the automatic evaluation pipeline are conducted on eight NVIDIA A800 GPUs.
For each evaluated system, we follow the official inference pipeline and adopt the highest-quality inference settings recommended by the provider, so that the comparison reflects each system under a strong default configuration rather than a resource-constrained setup.
For proprietary systems, we access their official APIs and use the highest-quality options available at the time of evaluation.

\subsection{Baseline Models}
\label{sec:models}

We evaluate 12 model settings, including both open-source and proprietary systems, and treat thinking-enabled variants as separate settings when they involve distinct inference procedures.
The open-source settings include Step1X-Edit and Step1X-Edit-thinking~\citep{liu2025step1x}, InstructPix2Pix~\citep{brooks2023instructpix2pix}, OmniGen2~\citep{wu2025omnigen2}, Bagel and Bagel-thinking~\citep{deng2025emerging}, FLUX.1-Kontext-dev~\citep{labs2025flux}, Qwen-Image-Edit and Qwen-Image-Edit-2511~\citep{wu2025qwen}, and FireRed-Image-Edit-1.1~\citep{qiao2026firered}.
The proprietary systems include GPT-image-1.5~\citep{openai2025gptimage15} and Nano Banana 2~\citep{google2026nanobanana2}.
Overall, this model suite covers early instruction-based image editing methods, recent open-source multimodal editing systems, and strong commercial black-box models, enabling us to examine whether multilingual degradation is consistent across different model families and deployment settings.

\subsection{Overall Results}

\begin{table*}[t]
\centering
\small
\setlength{\tabcolsep}{4pt}
\caption{Overall results on \benchmark{}. \textsc{SemAvg} is the mean of IF, TA, LSF, VC, and LP. EN and Non-EN denote semantic averages on the English and non-English subsets; Gap is their difference.}
\label{tab:overall-results}
\resizebox{\textwidth}{!}{
\begin{tabular}{l|cccccc|cccc|c@{\hspace{3pt}}c@{\hspace{3pt}}c}
\toprule
\multicolumn{1}{c|}{} & \multicolumn{6}{c|}{\textbf{Semantic}} & \multicolumn{4}{c|}{\textbf{Pixel}} & \multicolumn{3}{c}{\textbf{EN/Non-EN}} \\
\cmidrule(lr){2-7}\cmidrule(lr){8-11}\cmidrule(lr){12-14}
\textbf{Model} & \textbf{SemAvg}$\uparrow$ & \textbf{IF}$\uparrow$ & \textbf{TA}$\uparrow$ & \textbf{VC}$\uparrow$ & \textbf{LP}$\uparrow$ & \textbf{LSF}$\uparrow$ & \textbf{MSE}$\downarrow$ & \textbf{PSNR}$\uparrow$ & \textbf{SSIM}$\uparrow$ & \textbf{LPIPS}$\downarrow$ & \textbf{EN}$\uparrow$ & \textbf{Non-EN}$\uparrow$ & \textbf{Gap}$\downarrow$ \\
\midrule
Nano Banana 2 & \best{4.006} & \best{3.765} & \best{4.694} & \best{3.633} & \second{3.152} & \best{4.785} & \best{153.743} & \best{31.242} & \best{0.947} & \best{0.040} & \best{4.098} & \best{3.997} & \second{0.100} \\
GPT-image-1.5 & \second{3.577} & \second{3.114} & \second{4.065} & \second{3.456} & 2.752 & \second{4.499} & 1207.744 & 19.717 & 0.795 & 0.105 & \second{4.049} & \second{3.532} & 0.517 \\
Qwen-Image-Edit-2511 & 3.214 & 2.897 & 3.524 & 2.988 & 3.149 & 3.512 & 938.282 & 21.315 & 0.748 & 0.129 & 3.842 & 3.157 & 0.685 \\
FireRed-Image-Edit-1.1 & 3.204 & 2.933 & 3.857 & 2.976 & 2.612 & 3.642 & 1258.208 & 19.417 & 0.678 & 0.248 & 3.772 & 3.152 & 0.620 \\
Step1X-Edit & 2.802 & 2.327 & 2.701 & 2.625 & \best{3.432} & 2.924 & 937.224 & 27.150 & 0.902 & 0.057 & 3.358 & 2.751 & 0.606 \\
Bagel & 2.424 & 1.914 & 2.609 & 2.066 & 2.641 & 2.889 & 1049.571 & 25.036 & 0.913 & 0.070 & 2.986 & 2.373 & 0.614 \\
FLUX.1-Kontext-dev & 2.401 & 1.922 & 2.501 & 2.210 & 2.223 & 3.151 & 1697.211 & 17.888 & 0.701 & 0.164 & 3.110 & 2.337 & 0.773 \\
Bagel-thinking & 2.179 & 1.801 & 2.228 & 1.839 & 2.758 & 2.271 & 598.328 & 26.416 & \second{0.935} & \second{0.046} & 2.646 & 2.137 & 0.509 \\
Qwen-Image-Edit & 2.055 & 1.651 & 2.404 & 1.674 & 1.638 & 2.909 & 2742.375 & 15.828 & 0.677 & 0.231 & 2.477 & 2.017 & 0.460 \\
OmniGen2 & 1.734 & 1.272 & 2.137 & 1.324 & 1.551 & 2.383 & 5564.379 & 17.682 & 0.769 & 0.197 & 2.090 & 1.701 & 0.389 \\
Step1X-Edit-thinking & 1.604 & 1.054 & 1.174 & 1.177 & 1.522 & 3.091 & \second{513.660} & \second{28.403} & 0.918 & 0.046 & 2.122 & 1.557 & 0.565 \\
InstructPix2Pix & 1.190 & 0.892 & 1.623 & 0.448 & 0.574 & 2.411 & 4124.039 & 15.177 & 0.750 & 0.237 & 1.232 & 1.186 & \best{0.045} \\
\midrule
Average & 2.532 & 2.129 & 2.793 & 2.201 & 2.334 & 3.206 & 1732.064 & 22.106 & 0.811 & 0.131 & 2.982 & 2.491 & 0.490 \\
\bottomrule
\end{tabular}
}
\end{table*}

\begin{table*}[t]
\centering
\small
\setlength{\tabcolsep}{4pt}
\caption{Language coverage and cross-lingual degradation relative to English, aggregated across 12 model settings. Script, Dir., and Res.\ indicate writing system, text directionality, and resource level (H = High, MH = Mid-High, M = Mid, L = Low). \textsc{SemAvg} is the average semantic score, and \textbf{Std} denotes the standard deviation of model-wise $\Delta$Sem values.}
\label{tab:lang-degradation}
\resizebox{\textwidth}{!}{
\begin{tabular}{lccc|c|cccccc|c}
\toprule
\multicolumn{4}{c|}{\textbf{Meta}} & \multicolumn{1}{c|}{\textbf{Score}} & \multicolumn{6}{c|}{\textbf{Drop vs. EN}} & \multicolumn{1}{c}{\textbf{Std}} \\
\cmidrule(lr){1-4}\cmidrule(lr){5-5}\cmidrule(lr){6-11}\cmidrule(lr){12-12}
\textbf{Language} & \textbf{Script} & \textbf{Dir.} & \textbf{Res.} & \textbf{SemAvg}$\uparrow$ & \textbf{$\Delta$Sem}$\downarrow$ & \textbf{$\Delta$TA}$\downarrow$ & \textbf{$\Delta$LSF}$\downarrow$ & \textbf{$\Delta$IF}$\downarrow$ & \textbf{$\Delta$VC}$\downarrow$ & \textbf{$\Delta$LP}$\downarrow$ & \textbf{Std}$\downarrow$ \\
\midrule
English & Latn & LTR & H & \best{2.982} & 0.000 & 0.000 & 0.000 & 0.000 & 0.000 & 0.000 & 0.000 \\
Hebrew & Hebr & RTL & M & 2.126 & \best{0.856} & \best{1.168} & \best{1.551} & \second{0.675} & \second{0.565} & \second{0.319} & \best{0.461} \\
Arabic & Arab & RTL & MH & 2.202 & \second{0.780} & \second{1.005} & \second{1.258} & \best{0.685} & \best{0.615} & \best{0.337} & 0.397 \\
Bengali & Beng & LTR & M & 2.285 & 0.697 & 0.960 & 1.172 & 0.635 & 0.471 & 0.244 & \second{0.419} \\
Korean & Hang & LTR & MH & 2.438 & 0.543 & 0.717 & 0.911 & 0.513 & 0.336 & 0.241 & 0.289 \\
Russian & Cyrl & LTR & MH & 2.440 & 0.542 & 0.716 & 1.038 & 0.416 & 0.399 & 0.142 & 0.300 \\
Vietnamese & Latn & LTR & M & 2.540 & 0.442 & 0.514 & 0.941 & 0.301 & 0.339 & 0.113 & 0.195 \\
Yoruba & Latn & LTR & L & 2.549 & 0.433 & 0.477 & 1.028 & 0.286 & 0.247 & 0.128 & 0.168 \\
Japanese & Jpan & LTR & MH & 2.551 & 0.431 & 0.563 & 0.709 & 0.405 & 0.300 & 0.177 & 0.281 \\
Chinese & Hani & LTR & H & 2.677 & 0.304 & 0.363 & 0.387 & 0.320 & 0.282 & 0.171 & 0.374 \\
Spanish & Latn & LTR & H & 2.798 & 0.184 & 0.234 & 0.374 & 0.126 & 0.138 & 0.048 & 0.126 \\
Dutch & Latn & LTR & M & \second{2.800} & 0.181 & 0.212 & 0.326 & 0.155 & 0.153 & 0.060 & 0.117 \\
\midrule
Avg. (Non-EN) & -- & -- & -- & 2.491 & 0.490 & 0.630 & 0.881 & 0.411 & 0.350 & 0.180 & 0.284 \\
\bottomrule
\end{tabular}
}
\end{table*}

Table~\ref{tab:overall-results} reports the overall semantic and pixel-level performance of the evaluated systems on \benchmark{}.
We define \textsc{SemAvg} as the arithmetic mean of IF, TA, VC, LP, and LSF.

Nano Banana 2 achieves the strongest overall performance, ranking first on \textsc{SemAvg}, IF, TA, VC, and LSF while also leading all four pixel-level metrics; Step1X-Edit remains the best model on LP. Among open-source systems, Qwen-Image-Edit-2511 and FireRed-Image-Edit-1.1 are the strongest semantic performers, while Step1X-Edit and Step1X-Edit-thinking are especially competitive on layout preservation and pixel fidelity. Crucially, English performance exceeds the non-English average for every evaluated system, confirming that multilingual degradation is pervasive in current text-in-image editing models.\footnote{The English--non-English gap should be interpreted together with absolute performance: InstructPix2Pix has the smallest overall gap, but this mainly reflects similarly weak performance in both settings, whereas Nano Banana 2 combines the strongest absolute scores with the smallest gap among high-performing models.}

We further observe a recurring mismatch between semantic and pixel-level evaluation: Step1X-Edit-thinking and Bagel-thinking achieve excellent background fidelity but remain substantially weaker than the top models on semantic correctness, while Nano Banana 2 demonstrates that strong semantic editing and strong pixel preservation can co-exist in a top-tier multilingual system.

\subsection{Cross-Lingual Performance Degradation}
\label{sec:degradation}

Table~\ref{tab:lang-degradation} reports the language-level cross-lingual degradation results aggregated across the 12 evaluated model settings.
English is used as the reference language, and non-English languages are ordered by $\Delta$Sem in descending order.

Across our overall evaluation results, cross-lingual degradation is consistently observed across different models. Using English performance as the reference, Table~\ref{tab:lang-degradation} further shows that the degree of degradation is highly uneven across languages. Hebrew exhibits the largest semantic drop ($\Delta$Sem = 0.856), followed by Arabic (0.780); in contrast, Dutch (0.181) and Spanish (0.184) remain much closer to English. At the same time, the stability of degradation also varies substantially across languages. Hebrew shows the largest cross-model variance (\textbf{Std} = 0.461), whereas Dutch (0.117) and Spanish (0.126) are much more stable across models. Notably, Chinese also shows a relatively small degree of degradation (0.304). These findings suggest that the difficulty of cross-lingual text editing is not uniform across languages, and may be jointly related to factors such as writing direction, script characteristics, and language resource availability.

A clearer pattern emerges when examining the individual semantic dimensions separately. Averaged over all non-English languages, the largest drops relative to English are observed in text accuracy ($\Delta$TA = 0.630) and language/script fidelity ($\Delta$LSF = 0.881), whereas the declines in instruction following ($\Delta$IF = 0.411), visual consistency ($\Delta$VC = 0.350), and layout preservation ($\Delta$LP = 0.180) are noticeably smaller. This indicates that under multilingual conditions, current models are more likely to preserve the coarse-grained structure of the requested edit, yet struggle to generate the edited text correctly and faithfully in the target language. In other words, the main challenge lies not only in carrying out an edit operation, but also in producing text that is both linguistically correct and consistent with the target writing system.

\begin{figure*}[t]
\centering
\includegraphics[width=\textwidth]{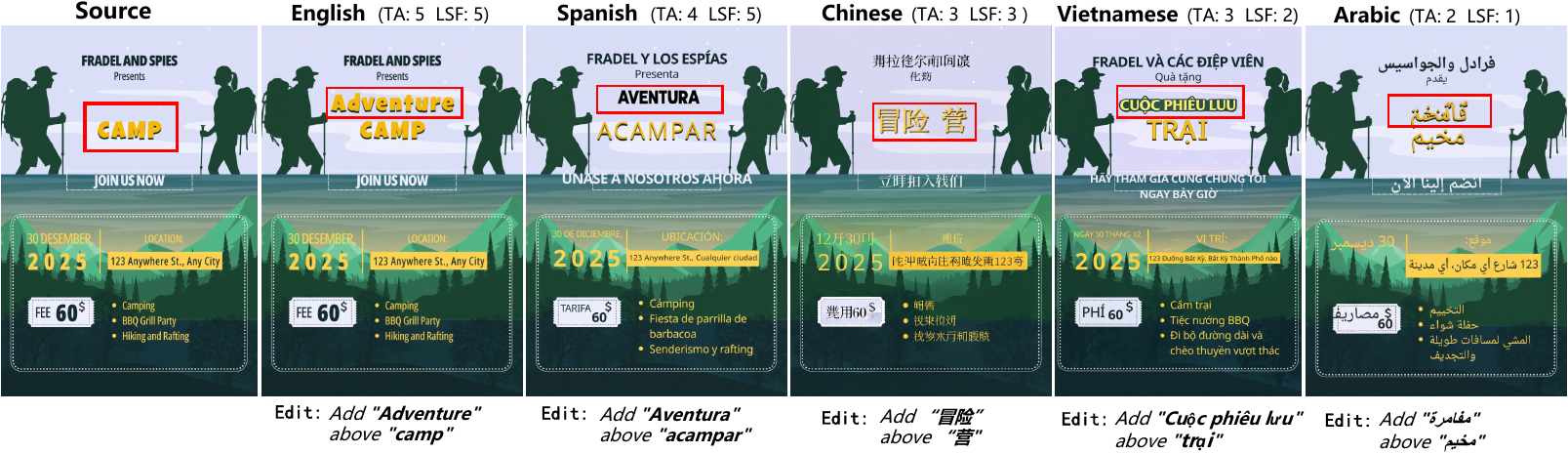}
\caption{Controlled cross-lingual comparison under the same edit template. Only the target text varies across languages, while semantic errors become more pronounced from Spanish to Arabic despite largely preserved layout.}
\label{fig:cross-lingual-case}
\end{figure*}

Figure~\ref{fig:cross-lingual-case} provides a direct qualitative view of this trend under a shared visual context. Relative to English and Spanish, Chinese and Vietnamese already show noticeable degradation in text accuracy or script fidelity, while Arabic exhibits a more severe failure in both dimensions despite preserving much of the global poster structure.

Overall, these results show that errors in multilingual text editing are concentrated more heavily in language-sensitive semantic dimensions than in coarse visual structure. 

\subsection{Language-Specific Failure Patterns}

\begin{figure}[t]
\centering
\includegraphics[width=\columnwidth]{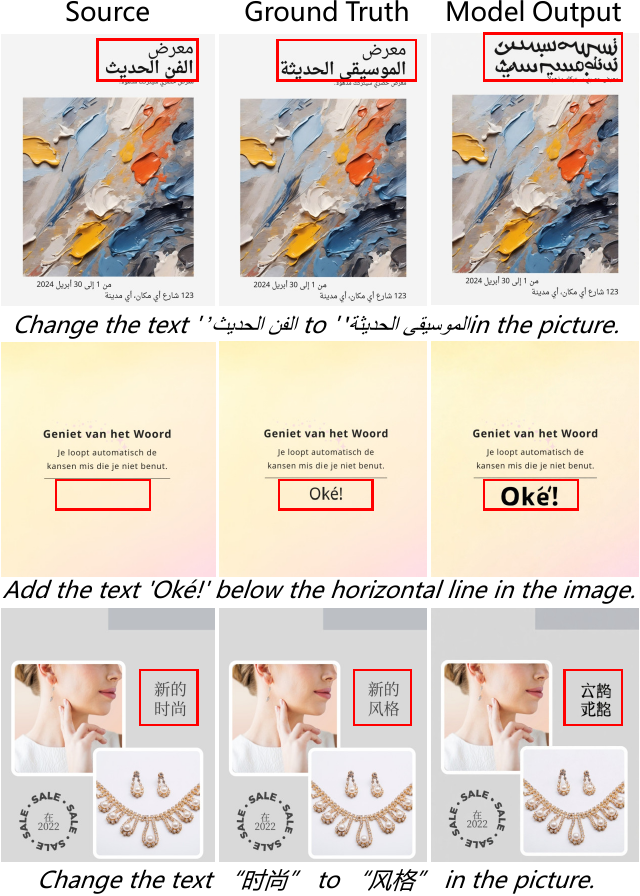}
\caption{Representative language-specific failure cases. From top to bottom: right-to-left, diacritic-related, and dense-character errors.}
\label{fig:language-specific-failures}
\end{figure}

The representative cases in Figure~\ref{fig:language-specific-failures} reveal that multilingual text editing errors are often language-specific rather than uniform across languages. In right-to-left scripts such as Hebrew and Arabic, models may preserve the overall layout while producing incorrect character order, unnatural directionality, or distorted script structure. In diacritic-sensitive languages such as Bengali, Vietnamese, and Yoruba, small glyph-level errors, including missing or corrupted diacritics, can substantially alter the target text despite limited pixel-level differences. In dense character systems such as Chinese, Japanese, and Korean, models may generate visually plausible characters that remain lexically incorrect. These cases suggest that multilingual editing failures arise not only from general image editing difficulty, but also from language-specific challenges at the script and character level, further supporting the inclusion of language/script fidelity in our evaluation framework.

\section{Conclusion}
In this paper, we introduced \benchmark{}, a benchmark for controlled evaluation of multilingual text-in-image editing. \benchmark{} contains 3,600 samples spanning 12 languages, 5 visual domains, and 7 editing operations, with human-edited references and region masks for fine-grained analysis. We further proposed a dual-track evaluation framework that combines LVM-based semantic assessment with mask-aware pixel-level metrics, enabling separate measurement of editing correctness and background preservation.

Experiments on 12 model settings show that multilingual degradation remains a persistent challenge for current systems. Across models, English consistently yields the strongest semantic performance, while non-English languages suffer larger drops, especially in text accuracy and language/script fidelity. We also identify a recurring \emph{semantic-pixel mismatch}: model outputs may preserve global layout and visual plausibility yet still fail to produce the correct edited text, particularly for scripts with directionality, dense characters, or diacritic-sensitive forms. These findings suggest that progress in multilingual text editing requires not only stronger generation quality, but also more language-aware modeling and evaluation. We hope \benchmark{} will serve as a useful testbed for developing more robust and more equitable text-in-image editing systems across languages.

\section*{Limitations}


\section*{Acknowledgments}

\bibliography{custom}

\appendix

\end{document}